# Deep learning model trained on mobile phone-acquired frozen section images effectively detects basal cell carcinoma

Junli Cao[1], B.S.; Junyan Wu[1], M.S.; Jing W. Zhang[2], M.D., Ph.D.; Jay J. Ye[3], M.D., Ph.D.; Limin Yu[4], M.D., M.S.

1. Meriti, Inc, 11400 National Blvd, Suite 143, Los Angeles, CA, 90064, USA
2. Carolinas Dermatology and Plastic Surgery, 1706 Saint Julian Place, Columbia, SC, 29204, USA
3. Dahl-Chase Pathology Associates, 417 State St., Suite 540, Bangor, ME 04401, USA
4. Department of Pathology, Beaumont Health System, Royal Oak, MI, USA

Junli Cao: evejlcao@ucla.edu

Junyan Wu: wujy1208@gmail.com

Jing W. Zhang: caroderm@gmail.com

Jay J. Ye: jye@dahlchase.com

Limin Yu: limin.yu@beaumont.org

**Abstract**

**Background**: Margin assessment of basal cell carcinoma using the frozen section is a common task of pathology intraoperative consultation. Although frequently straightforward, the determination of the presence or absence of basal cell carcinoma on the tissue sections can sometimes be challenging. We explore if a deep learning model trained on mobile phone-acquired frozen section images can have adequate performance for future deployment. **Materials and Methods**: One thousand two hundred and forty-one (1241) images of frozen sections performed for basal cell carcinoma margin status were acquired using mobile phones. The photos were taken at 100x magnification (10x objective). The images were downscaled from a 4032 x 3024 pixel resolution to 576 x 432 pixel resolution. Semantic segmentation algorithm Deeplab V3 with Xception backbone was used for model training. **Results:** The model uses an image as input and produces a 2-dimensional black and white output of prediction of the same dimension; the areas determined to be basal cell carcinoma were displayed with white color, in a black background. Any output with the number of white pixels exceeding 0.5% of the total number of pixels is deemed positive for basal cell carcinoma. On the test set, the model achieves area under curve of 0.99 for receiver operator curve and 0.97 for precision-recall curve at the pixel level. The accuracy of classification at the slide level is 96%. **Conclusions**: The deep learning model trained with mobile phone images shows satisfactory performance characteristics, and thus demonstrates the potential for deploying as a mobile phone app to assist in frozen section interpretation in real time.

**Introduction**

Basal cell carcinoma is the most common skin cancer worldwide and continuing to increase in incidence [1]. Surgical excision is a common modality of treatment and the frozen section can be judiciously used intraoperatively to determine the margin status of the excision, particularly for the lesions in the critical anatomical sites such as the nose, cheeks, eyelids, chin, lips and forehead [2]. For these locations, a persistent/recurrent disease from an incomplete primary excision would demand a far more aggressive and disabling secondary surgery [2].

Deep learning models trained with microscopic images have been used to detect, classify and grade cancers with promising results [3-12]. Most of the studies used digitized images obtained from the permanent sections derived from formalin-fixed and paraffin-embedded tissue, acquired using the whole slide imaging technique.

Although the interpretation of the presence or absence of basal cell carcinoma on the frozen section slides are straightforward most of the time, it can sometimes be challenging. We seek to explore if a deep learning model trained on mobile phone-acquired frozen section images can have adequate performance for its future deployment as a mobile phone app for real-time assistance to the pathologists.

**Materials and methods**

One thousand two hundred and forty-one (1241) images of frozen sections performed for basal cell carcinoma margin status was acquired using two different mobile phones by two dermatopathologists (JWZ and JJY) from their institutions. The photos were taken at 100x magnification (10x objective). Five hundred ninety-eight (598) images contain basal cell carcinoma and 643 images are negative for basal cell carcinoma.

These images were randomly partitioned into the training set (1,116 images) and the test set (125 images), with each set containing a similar proportion of images with or without basal cell carcinoma.

The images were downscaled from a 4032 x 3024 pixel resolution to 576 x 432 pixel resolution.

Semantic segmentation algorithm Deeplab V3 [12] with Xception [13] backbone developed by Google was used to train our model. The weights were pre-trained on PASCAL VOC 2012 and fine-tuned with our own dataset. In this study, the Atrous Spatial Pyramid Pooling (ASPP) rate was (6, 12, 18) and the output stride was 16. We used Adam optimizer with an initial learning rate of 1E-3 and sigma of 1E-8.

The performance of the trained model was evaluated by the following metrics.

Mean IOU (Jaccard Index) is the average IOU over all classes. The IOU for each class was defined as follows:

IOU = TP / (TP + FP + FN)

To further evaluate our segmentation model, we also reported the pixel level receiver operating characteristic curve (ROC) and precision-recall (PR) curve.

Image level basal cell carcinoma classification was performed by calculating the number of positive pixels in each model outputs with respect to the input image. A threshold of 0.5% was used.

## Results

Our deep learning model on basal cell carcinoma achieved satisfactory results. Figure 1(A-E) showed the original image, our model classification and ground-truth labeling respectively from the leftmost to the rightmost for selected examples. As shown, our model prediction identified the overall shapes and locations of the lesions in most images in the test set. In some locations the prediction could be coarse, which might have resulted from the coarse ground-truth labels in our dataset. Figure 1A showed a high-quality segmentation output that largely overlapped with the ground-truth label. Figure 1E showed a representative image with false positive prediction. The possible reason for this might be because the convolution neural networks were less sophisticated in catching the context information around the normal tissues, and the context information was sometimes crucial in medical images.

Our model achieved 0.96 IOU for the normal pixels and 0.801 for the cancer pixels, with a mean IOU of 0.881.

We then extended our basal cell carcinoma segmentation model to slide level basal cell carcinoma classification model by calculating the number of predicted positive pixels with respect to the input image. If the number of predicted positive pixels exceeded 0.5% of the total pixels of an image, we would classify the image as positive for basal cell carcinoma. We obtained 96% accuracy of the slide level classification. Specifically, 65/65 images with basal cell carcinoma were correctly detected (100% sensitivity) and 55/60 tumor-free images were correctly classified, with 91.7% specificity.

Furthermore, because the segmentation task was essentially a pixel level classification task, we provided pixel level classification analysis in this study. We mainly analyzed the model by ROC and PR curves. As shown in Figure 2, our model generated a near-perfect ROC curve with AUC= 0.99. Figure 3 showed a PR curve of our model with ISO-F1 curve on the top. The PR curve described the trade-off between the positive predicted value (precision) and true positive rate (recall) of a binary classification using different probability threshold. Similarly, we obtained 0.97 AUC for the PR curve despite that PR curve was a relatively stringent measure for the classification task.

# Discussion

This study has been designed to assess if deep learning models trained with digital images of frozen section slides acquired using mobile phones can have satisfactory performance so as to be deployed as a mobile app.

A deep learning model based on the semantic segmentation was chosen. We employ the state-of-art Deeplab v3[12] network with Xception[13] backbone for our segmentation task. Instead of using regular convolution layers, the Deeplab v3 adopts the astrous convolution layers, which provides a wider field of view but at the same computational costs. In a similar fashion, the Xception network uses depthwise separate convolution layers. The advantage of this convolution is that it first performs a spatial convolution then follow by a depthwise convolution so that it has fewer parameters and thus it requires fewer operations to compute.

The decision to downscale the images was due to the following two considerations. First, when pathologists make the diagnosis of basal cell carcinoma, the judgement relies more on low-magnification features than on high magnification features. Second, downscaling may make the feature mobile app run faster.

There are two factors that may potentially adversely affect the performance of the model. First, it is reasonable to assume that images acquired by a mobile phone are of less quality in comparison to images acquired by the whole slide imaging technique. Second, the image quality of frozen section slides viewed through a microscope is in general less optimal than that of the permanent section slides sectioned from formalin-fixed and paraffin-embedded tissue.

Despite these inherent disadvantages, the accuracy of classification at the slide level is 96%. While the false positive rate is 9%, the false negative rate is 0%. In additions, the images were obtained by two different dermatopathologists at two different institutions, using different mobile phones. The staining characters of the frozen sections slides may vary from institution to institution. Taking all these into considerations, an accuracy rate of 96% seems to be quite satisfactory. It is conceivable that a model trained with more images and images from additional institution may improve both the accuracy and robustness of the model.

The output of the model delineates the area on the slide that is interpreted by as basal cell carcinoma. A pathologist can then further scrutinize the area to ensure that basal cell carcinoma is not missed in these areas.

In conclusion, the results of our feasibility study indicate that the deep learning model built on frozen section images acquired using a mobile phone can potentially be deployed as a mobile app to provide real-time assistance to the pathologists interpreting frozen section slides for the evaluation of the margin status of basal cell carcinoma excision.

Funding: This research did not receive any specific grant from funding agencies in the public, commercial, or not-for-profit sectors.

**Reference:**

1. Cameron MC, Lee E, Hibler BP, Barker CA, Mori S, Cordova M, Nehal KS, Rossi AM. Basal cell carcinoma: Epidemiology; pathophysiology; clinical and histological subtypes; and disease associations. J Am Acad Dermatol 2019;80(2): 303-817.

2. Nicoletti G, Brenta F, Malovini A , Musumarra G, Scevola S, Faga A. Study to determine whether intraoperative frozen section biopsy improves surgical treatment of non-melanoma skin cancer. Mol Clin Oncol. 2013; 1(2): 390–394.

3. Janowczyk A, Madabhushi A. Deep learning for digital pathology image analysis: a comprehensive tutorial with selected use cases. J Pathol Inform. 2016; 7(1):29.

4. Djuric U, Zadeh G, Aldape K, Diamandis P. Precision histology: how deep learning is poised to revitalize histomorphology for personalized cancer care. NPJ Precis Oncol. 2017;1(1):22

5. Olsen TG, Jackson B H, Feeser TA, Kent MN, Moad JC, Krishnamurthy S, Lunsford DD, Soans RE. Diagnostic performance of deep learning algorithms applied to three common diagnoses in dermatopathology. J Pathol Inform 2018;9:32.

6. Liu Y, Kohlberger T, Norouzi M, Dahl GE, Smith JL, Mohtashamian A, Olson N, Peng L, Hipp JD, Stumpe MC. Artificial Intelligence–Based Breast Cancer Nodal Metastasis Detection. Archives of Pathology & Laboratory Medicine 2019; 143 (7): 859–868.

7. Steiner DF, MacDonald R, Liu Y, Truszkowski P, Hipp JD, Gammage C, Thng F, Peng L, Stumpe MC. Impact of Deep Learning Assistance on the Histopathologic Review of Lymph Nodes for Metastatic Breast Cancer. Am J Surg Pathol. 2018;42(12):1636-1646.

8. Sanyal P, Mukherjee T, Barui S, Das A, Gangopadhyay P. Artificial intelligence in cytopathology: A neural network to identify papillary carcinoma on thyroid fine-

9. Xu L, Walker B, Liang PI, Tong Y, Xu C, Su YC, Karsan A. Colorectal cancer detection based on deep learning. J Pathol Inform 2020;11:28

10. Parwani AV. Commentary: Automated diagnosis and gleason grading of prostate cancer – are artificial intelligence systems ready for prime time?. J Pathol Inform 2019;10:41

11. Nir G, Hor S, Karimi D, Fazli L, Skinnider BF, Tavassoli P, et al. Automatic grading of prostate cancer in digitized histopathology images: Learning from multiple experts. Med Image Anal 2018;50:167-80

12. Liang-Chieh Chen, George Papandreou, Iasonas Kokkinos, Kevin Murphy, Alan L Yuille. Deeplab: Semantic image segmentation with deep convolutional nets, atrous convolution, and fully connected crfs. TPAMI, 2017.

13. Chollet, F. Xception: Deep learning with depthwise separable convolutions. In: CVPR. (2017)

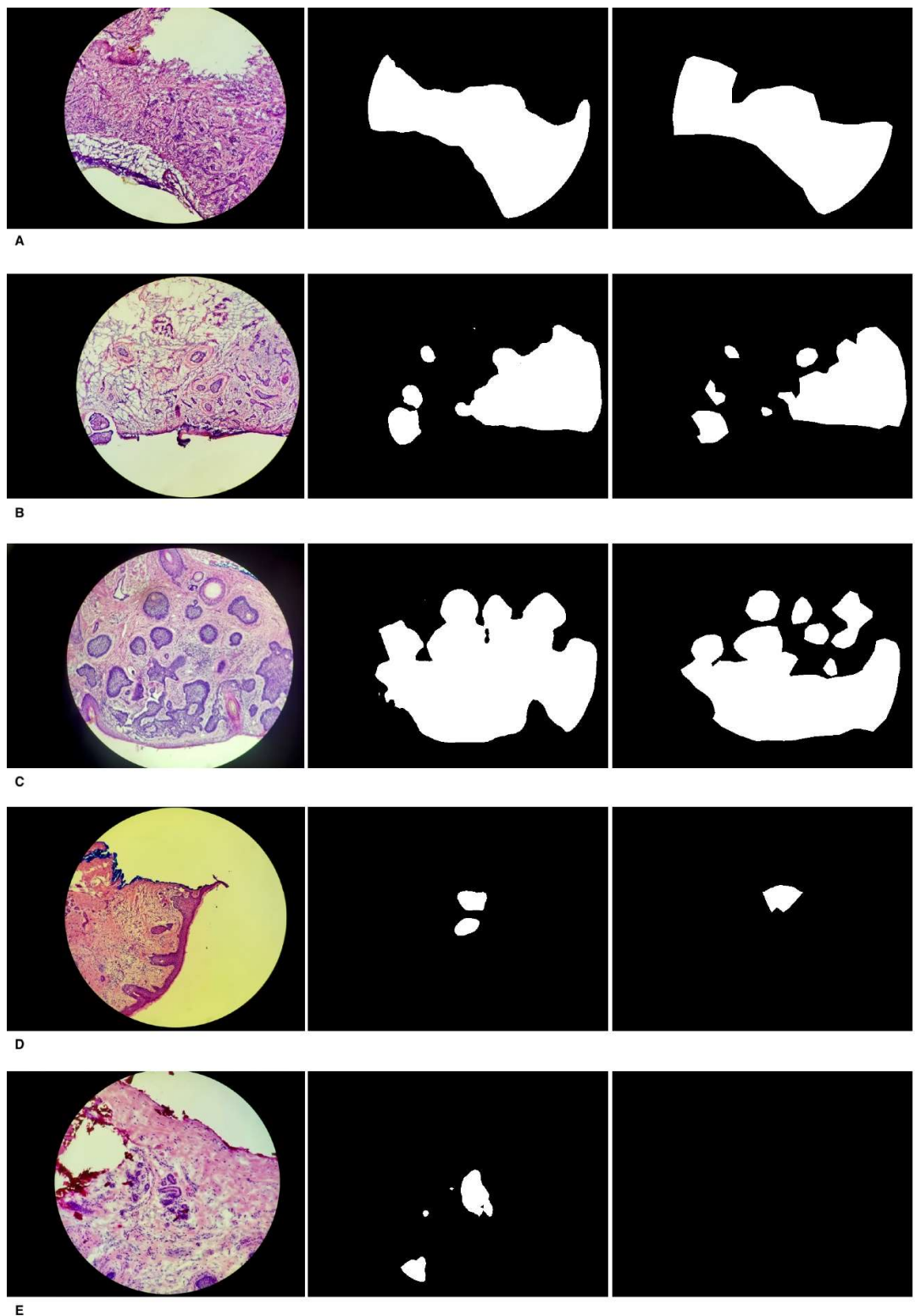

Figure 1. Segmentation of basal cell carcinoma. The left column is microscopic images. The middle column is our model predictions of basal cell carcinoma. The right column is the ground-truth labels.

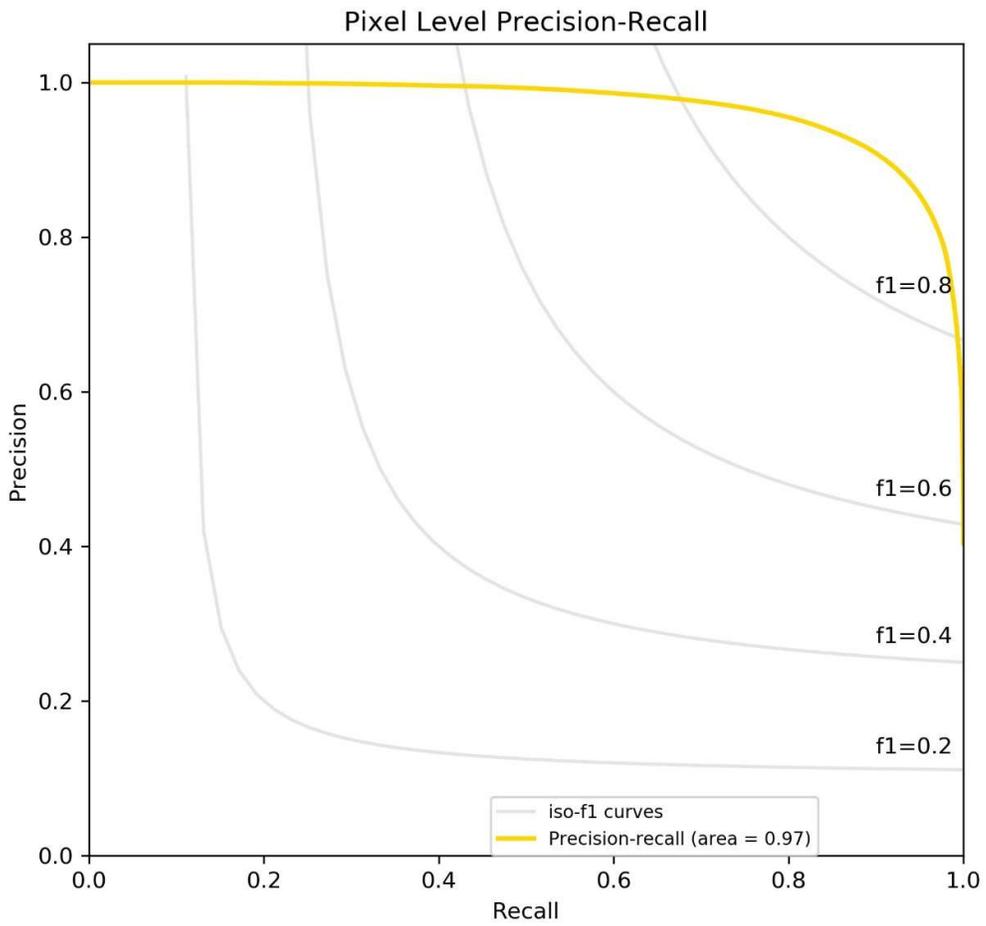

Figure 2. Receiver operating characteristic curve of the pixel level classification model. The area under the curve (AUC) of the classification model is 0.99.

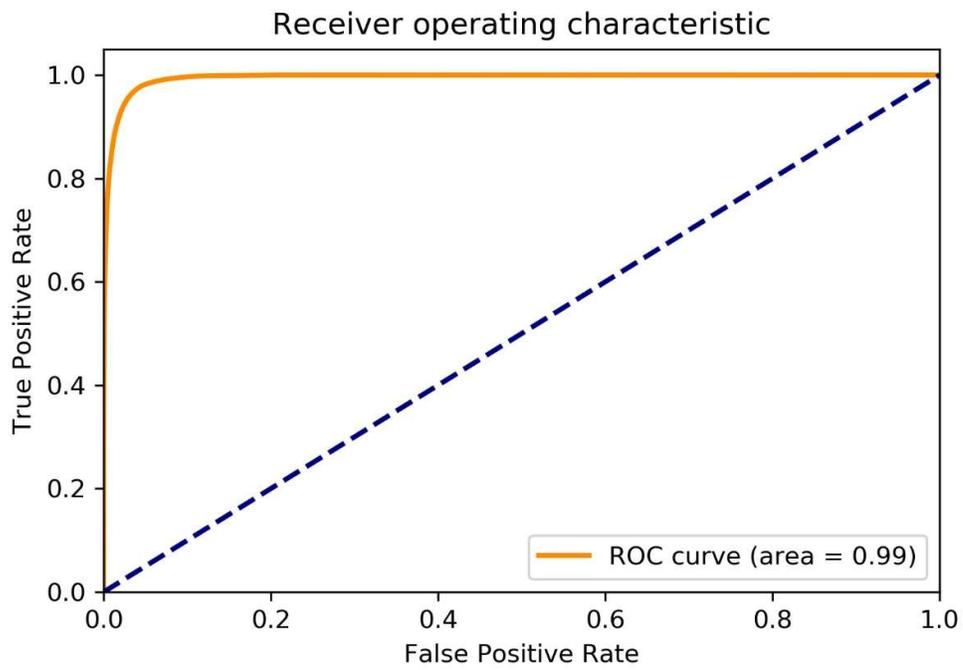

Figure 3. Precision-recall curve of the pixel level classification model with ISO-F1 curve on the top. The area under the curve (AUC) of the classification model is 0.97.